%% file: main.tex
\documentclass{article}
\usepackage{subfiles}
\usepackage{natbib}
\usepackage{xcolor}

\usepackage{arxiv}

\usepackage[utf8]{inputenc} 
\usepackage[T1]{fontenc}    
\usepackage{hyperref}       
\usepackage{url}            
\usepackage{booktabs}       
\usepackage{amsfonts}       
\usepackage{nicefrac}       
\usepackage{microtype}      
\usepackage{graphicx}
\usepackage{subcaption}
\usepackage{amsmath}
\usepackage{multirow}
\usepackage[normalem]{ulem}
\graphicspath{ {./img/} }

\title{Parameter‐Efficient Multimodal Instruction Tuning for Romanian Vision–Language Models}

\author{
George-Andrei Dima, Răzvan-Alexandru Smădu, and Dumitru-Clementin Cercel\thanks{Corresponding author.} \\
  National University of Science and Technology POLITEHNICA Bucharest, Bucharest, Romania\\
  \texttt{\{george\_andrei.dima,razvan.smadu\}@stud.acs.upb.ro,dumitru.cercel@upb.ro} \\
}

\newcommand{\accessDate}{20 June 2026}

\newcommand{\flickrRO}{RoFlickr30k-Caption}
\newcommand{\flickrROQA}{RoFlickr30k-QA}

\newcommand{\llamaBase}{LLaMA 3.2 11B Vision}
\newcommand{\llamaRO}{LLaMA 3.2 11B-Vision-RoVQA}

\newcommand{\llavaBase}{LLaVA-v1.6-Mistral-7B}
\newcommand{\llavaRO}{LLaVA-v1.6-Mistral-7B-RoVQA}

\newcommand{\qwenBase}{Qwen2-VL-7B-Instruct}
\newcommand{\qwenRO}{Qwen2-VL-RoVQA}

\begin{document}
\maketitle
\begin{abstract}

Focusing on low-resource languages is an essential step toward democratizing generative AI. In this work, we contribute to reducing the multimodal NLP resource gap for Romanian. We translate the widely known Flickr30K dataset into Romanian and further extend it for visual question answering by leveraging open-source LLMs. We demonstrate the usefulness of our datasets by fine-tuning open-source VLMs on Romanian visual question answering. We select VLMs from three widely used model families: LLaMA 3.2, LLaVA 1.6, and Qwen2. For fine-tuning, we employ the parameter-efficient LoRA method. Our models show improved Romanian capabilities in visual QA, as well as on tasks they were not trained on, such as Romanian image description generation. The seven-billion-parameter \qwenRO{} obtains top scores on both tasks, with improvements of +2.29\% and +4.45\% in BERTScore F\textsubscript{1} on VQA and captioning, respectively, over its original version. Finally, the models show substantial reductions in grammatical errors compared to their original forms, indicating improvements not only in language understanding but also in Romanian fluency.

\end{abstract}

\keywords{Visual Language Models \and Romanian \and PEFT}

\section{Introduction}
\subfile{1introduction}

\section{Related Work}
\label{sec:related_work}
\subfile{2relatedworkplus}

\section{Methodology}
\label{sec:methodology}
\subfile{3method}

\section{Experiments and Results}
\label{sec:results}
\subfile{4results}

\section{Conclusions}
\label{sec:conclusions}

\subfile{5conclusion}

\bibliographystyle{unsrtnat}
\bibliography{references}

\appendix
\subfile{appendix}

\end{document}

%% file: 1introduction.tex
Recent advances in vision-language modeling (VLM) have enabled substantial progress on various multimodal tasks, including image captioning and visual question answering.  However, the vast majority of these models and benchmarks remain centered on English or other high-resource languages, limiting their applicability in low-resource settings.  In particular, Romanian lacks both large-scale multimodal corpora and task-specific instruction-tuned models, hindering progress in downstream applications that require joint visual and linguistic understanding.

Prior work has begun to adapt multimodal large language models (MLLMs) to underrepresented languages through dataset translation, parameter-efficient fine-tuning, and synthetic instruction generation.  
LLaVA-NDiNO for Italian \citep{musacchio2024llava}, LaVy for Vietnamese \citep{tran2024lavy}, and X-LLaVA for Korean and Chinese \citep{shin-etal-2024-x} are only a few of the works that report gains by translating widely used datasets from the English language and by using LLM-generated instructions. LRM-LLaVA \citep{li2025lrm} and Maya \citep{alam2024maya} extend this approach to multilingual settings.
To our knowledge, no previous work has conducted a systematic investigation of multimodal instruction tuning for Romanian.

In this work, we take a first step towards closing this gap.  
We construct \textbf{\flickrRO{}}, a translation of the English Flickr30K \citep{young2014image} captions into Romanian, and extend it with automatically generated question-answer (QA) pairs to form \textbf{\flickrROQA{}}.  
The resulting corpus provides almost 160,000 Romanian captions and 32,000 aligned visual QA (VQA) pairs.
Each caption was translated and proofread by native speakers, ensuring grammatical fidelity and stylistic consistency.  
The question-answer pairs are produced by a 70-billion-parameter LLaMA 3.3 \citep{grattafiori2024llama3herdmodels} model conditioned solely on textual descriptions, thus avoiding image hallucination. The VQA test split was additionally validated by a human annotator, who revised questions and answers where necessary.

To establish strong baselines, we perform parameter-efficient adaptation of three state-of-the-art vision-language architectures:
\llamaBase{} \citep{grattafiori2024llama3herdmodels}, \qwenBase{} \citep{wang2024qwen2}, and \llavaBase{} \citep{liu2024llavanext}.
We fine-tune only the parameters in the low-rank adaptation (LoRA) \citep{hu2022lora} modules on the Romanian VQA corpus, preserving the pretrained vision encoders and significantly reducing compute requirements.  Our experiments demonstrate consistent gains in both VQA performance and image captioning quality after adaptation. 
Our strongest VQA model, \qwenRO{}, achieves a BERTScore~F\textsubscript{1} of 77.38\% and a ROUGE-L~F\textsubscript{1} of 53.54\%, outperforming its original counterpart by 2.29\% and 8.40\%, respectively.  These results confirm that a predominantly English model can benefit from a relatively small amount of Romanian supervision. In image captioning, the adapters generalize without additional task-specific training. \qwenRO{} obtains the highest BERTScore~F\textsubscript{1}, while \llavaRO{} achieves the highest ROUGE-L~F\textsubscript{1} and BLEU score, with a BLEU improvement of 12.76 points over its base model.

We further quantify improvements in grammatical fluency using an automatic LLM-as-judge protocol \citep{zheng2023judge}, showing how Romanian instruction tuning influences surface-level grammatical errors. We evaluate hallucinations under image perturbations, finding that fine-tuning reduces hallucinations in the original VQA setting. Finally, our work aims to answer the four research questions:

\begin{itemize}

    \item \textbf{RQ1}: Can LLMs be effectively used to generate valuable training data for low-resource languages, enabling smaller models to bootstrap their language capabilities?
    
    \item \textbf{RQ2}: Do parameter-efficient fine-tuning methods enhance the low-resource language capabilities of English-centric vision-language models?

    \item \textbf{RQ3}: How grammatically correct are the answers produced by the LLMs in Romanian?

    \item \textbf{RQ4}: How resilient are the models to hallucinations against perturbed inputs?

\end{itemize}

Our main contributions are:
\begin{itemize}

\item We release \flickrRO{}, a Romanian translation of the Flickr30K captions, and \flickrROQA{}, the first Romanian visual QA corpus \footnote{ The complete dataset, including Romanian VQA annotations, Romanian captions, images, and image identifiers, is avaialable at\url{https://huggingface.co/datasets/GRAI-UNSTPB/Flickr30K-RoQA-v1}, accessed on \accessDate{}.}.

\item We present a lightweight instruction-tuning pipeline that brings task-specific Romanian knowledge into existing MLLMs without expensive full-model training. We made the resulting LoRA adapters publicly available to facilitate future work on Romanian multimodal understanding\footnote{LoRA adapters for \llamaRO{}, \llavaRO{}, and \qwenRO{} are available at \url{https://huggingface.co/collections/GRAI-UNSTPB/romanian-vision-language-models}, accessed on \accessDate{}.}.

\item We run comprehensive experiments on Romanian visual question answering and image captioning, together with a grammatical-error analysis, hallucination behavior, and robustness under perturbations.

\end{itemize}

The remainder of this paper is organized as follows.  \S\ref{sec:related_work} reviews related work on vision-language models for low-resource languages, multilingual multimodal corpora, benchmarks, and synthetic data generation.  \S\ref{sec:methodology} describes our dataset construction, model selection, and fine-tuning procedure.  \S\ref{sec:results} presents quantitative results for both Romanian VQA and image captioning, as well as a grammatical error analysis. \S\ref{sec:discussions} presents our discussion addressing the research questions.  Finally, \S\ref{sec:conclusions} concludes and outlines directions for future research.

%% file: 2relatedworkplus.tex
\subsection{Base Vision-Language Models}

Core vision-language model architectures represent the foundation of the recent breakthroughs in the literature, such as the vision transformer (ViT) \citep{dosovitskiy2021an}, LLaVA and its variants \citep{liu2023llava,liu2023improvedllava,liu2024llavanext}, and two integrated VLM families, LLaMA Vision \citep{grattafiori2024llama3herdmodels} and Qwen-Vision \citep{wang2024qwen2}.
These architectures merge textual and visual features into unified, valuable representations that enable learning meaningful outputs. 
These are the same backbones we evaluate in this paper.

\textbf{ViT.} The vision transformer architecture was introduced by \citet{dosovitskiy2021an} to adapt the transformer model for visual tasks and step away from existing convolutional networks-based solutions. The key contribution is a novel method for mapping images into visual tokens. Images are split into patches, which are linearly projected into the token embedding space and then processed by a standard transformer encoder. This method achieves results comparable to or exceeding those in image classification.

\textbf{LLaVA.} The LLaVA architecture, introduced by \citet{liu2023llava}, is composed of three modules: a vision encoder, a projection module, and a language model. In the original LLaVA, the vision encoder is a ViT backbone, which encodes images into a latent representation. These image embeddings are then processed by the projection module, i.e., a linear projection layer, that maps them into a sequence of visual tokens. Finally, the visual tokens are concatenated with text tokens, and the combined sequence is fed to the LLM to generate the answer. The LLaVA originally used the CLIP ViT-L/14 \citep{radford2021learningtransferablevisualmodels} as vision encoder and Vicuna \citep{vicuna2023} as language model.
\textbf{LLaVA-1.5} \citep{liu2023improvedllava} showed improved results with simple upgrades to the LLaVA architecture. The authors replaced the vision encoder with CLIP-ViT-L/336px and increased the depth of the projection module by turning it into a multi-layer perceptron.
\textbf{LLaVA-NeXT} \citep{liu2024llavanext} improves the model’s resolution capabilities while also expanding the set of supported LLM backbones.

\textbf{Qwen2 VL.} Qwen2-VL-7B-Instruct \citep{wang2024qwen2} follows the LLaVA architecture, using a vision transformer as the encoder and Qwen2 7B as the language model. It introduces two enhancements for merging image and text representations: naive dynamic resolution, which enables processing of images at arbitrary resolutions, and multimodal rotary position embedding, which encodes positional information from multimodal features.  

\textbf{LLaMA Vision.} \llamaBase{} \citep{grattafiori2024llama3herdmodels} also employs a LLaVA architecture with small modifications. Similar to LLaVA, the input image is processed by a vision transformer. Instead of projecting image embeddings into the token space, embeddings are integrated via an image adapter. The image adapter consists of cross-attention layers that connect the image representations and the representations produced by the language model. These layers are inserted after every four self-attention layers in the language model.

\subsection{Adaptation of Vision-Language Models}

Building on the existing base models, the community has introduced numerous vision-language models fine-tuned for a wide array of tasks and domains, as shown in Table~\ref{tab:vlm_specific_tasks}. This growth is driven by the availability of high-quality synthetic datasets generated with recent large language models.

\begin{table}[ht]
\centering
\caption{Vision--Language Models Fine-Tuned for Specific Tasks/Domains.}
\label{tab:vlm_specific_tasks}
\begin{tabular}{p{2.5cm}p{2.5cm}p{3cm}p{3cm}p{2.5cm}}
\hline
\addlinespace
\textbf{Model} & \textbf{Architecture} & \textbf{Data Strategy} & \textbf{Training Strategy} & \textbf{Domain}\\
\addlinespace
\hline
\addlinespace

LLaVA-Med (2023) &
LLaVA&
image-caption pairs from PMC, augmented with machine-generated instructions &
projector + LLM fine-tuning &
Biomedicine (medical images) \\
\addlinespace

Quilt-LLaVA (2023) &
LLaVA &
machine-generated Q/A pairs from histopathology data &
projector tuning + instruction fine-tuning&
Histopathology \\
\addlinespace

Math-LLaVA (2024) &
LLaVA-1.5 &
40K original math QA samples, augmented with 320K machine-generated QA samples&
full fine-tuning &
Mathematics\\
\addlinespace

PathChat (2024) &
similar to LLaVA-1.5, with custom weights for the vision encoder &
real pathology data refined via machine-assisted curation &
vision encoder pretraining + full instruction fine-tuning &
Histopathology \\
\addlinespace

G-LLaVA (2025) &
LLaVA &
machine-generated descriptions and QA pairs built on existing geometry datasets &
full fine-tuning &
Geometry\\
\addlinespace

MedVP-LLaVA (2025) &
LLaVA &
machine augmented medical VQA datasets &
full fine-tuning &
Medical VQA\\
\addlinespace

\hline
\end{tabular}
\end{table}

\textbf{Medical Domain.}
VLMs are particularly useful in the medical domain, where most of the tasks include a mix of visual and textual information. LLaVA-Med \citep{li2023llavamed} focused on biomedicine and addresses the problem of medical visual question answering. The authors based their model on the LLaVA architecture and performed a two-step fine-tuning: medical concept alignment followed by medical instruction tuning. The alignment stage aimed to train the projection layer to better map the biomedical features. The instruction tuning stage additionally tuned the LLM to better converse in the biomedical context. For the training data, the authors leveraged image-caption pairs taken from PMC. For the second stage of fine-tuning, the captions were synthetically transformed into instruction-response pairs. They further validated their method on standard medical VQA datasets, achieving strong performance.

MedVP-LLaVA \citep{zhu-etal-2025-guiding} aimed to address the shortcomings of VLMs in the medical VQA task by visual grounding, i.e., directing the model to focus on specific areas of the medical image. The authors generated a dataset with markers for regions of interest through a two-step process: automatic extraction of entities from captions and automatic object detection of the extracted entities in the image. All model components were fine-tuned, and the resulting model was validated on medical VQA datasets, outperforming the state of the art.

Histopathology is another medical field where VLMs have been tested. Quilt-LLaVA \citep{seyfioglu2023quiltllava} aimed to address visual question answering on histopathology images. The authors generate instructions synthetically from histopathology training video transcripts. They followed a two-stage fine-tuning method similar to LLaVA-Med: first, fine-tuning with an alignment step, where only the MLP projector is trained, followed by an instruction-tuning step where only the vision encoder remains frozen. The authors reported significant improvements over the state of the art on medical VQA tasks, demonstrating the efficacy of synthetic data generation.

PathChat \citep{lu2024pathchat} is a chat-based assistant for histopathology built on top of LLaVA-1.5. The authors replaced the general vision encoder weights with the domain-specific encoder UNI \citep{Chen2024UNI}. Fine-tuning was performed in two stages: further training the image encoder with image-caption pairs taken from the CONCH corpus \citep{Lu2024CONCH}, followed by full instruction tuning. Regarding the instruction dataset, LLMs were leveraged to structure the source text and assist human annotators in creating the instructions. PathChat achieved state-of-the-art results for multiple-choice question-answering tasks, further demonstrating that domain-specific models can outperform significantly larger general-purpose models.

\textbf{Mathematical Reasoning.}
Another domain where VLMs have proven useful is multimodal mathematical reasoning. Math-LLaVA \citep{shi-etal-2024-math} was trained using a mix of original and synthetic data: the authors collected text-image pairs from existing mathematical datasets and further generated instructions and responses using GPT-4V. They trained two Math-LLaVA versions, one using only the original data and another including the additional synthetic data, and demonstrated that leveraging large language models for data generation can significantly improve model performance. Finally, the authors reported results notably better than the base model and comparable to those of closed-source VLMs.

G-LLaVA \citet{gao2025gllava} focuses on exploring the capabilities of VLMs in solving geometry problems. The authors leveraged LLMs to generate figure captions from geometric symbolic text and further construct a multimodal instruction dataset. Training G-LLaVA employed a two-phase approach: an alignment stage, involving fine-tuning on geometric figures and their captions, followed by geometric instruction tuning. The authors noted that freezing the LLM during alignment yields better performance and fewer hallucinations. Finally, G-LLaVA achieved state-of-the-art results on geometry tasks, outperforming much larger models such as GPT-4V.

\subsection{Vision-Language Models for Low-Resource Languages}

\begin{table}[ht]
\centering
\caption{Vision--Language Models for Low‑Resource Languages.}
\label{tab:vlm_low_resource}
\begin{tabular}{p{2.5cm}p{2.5cm}p{3cm}p{3cm}p{2.5cm}}
\hline
\addlinespace
\textbf{Model} & \textbf{Architecture} & \textbf{Data Strategy} & \textbf{Training Strategy} & \textbf{Languages}\\
\addlinespace
\hline
\addlinespace
LLaVA-NDiNO (2024) & 
LLaVA-NeXT & 
mix of Italian-native and machine translated & 
full training & 
Italian \\
\addlinespace

LaVy (2024) & 
LLaVA-1.5 (with Vistral-7B as Language Model) & 
machine translated then refined & 
projector pretraining + instruction fine-tuning & 
Vietnamese \\
\addlinespace

X-LLaVA (2024) & 
LLaVA-1.5 & 
synthetic (generated) & 
LLM vocab expansion + multimodal fine-tuning & 
Korean\\
\addlinespace

PALO (2025) & 
LLaVA-1.5 &  
machine translated & 
adapter pre-training + instruction fine-tuning& 
10 languages \\
\addlinespace

Maya (2024) & 
LLaVA (SigLIP vision encoder + Cohere Aya-23 8B LLM) & 
machine translated (hybrid pipeline for 7 langs) & 
vision-language alignment + multilingual SFT & 
10 languages (8 in Maya's original dataset + 2 covered in PALO's dataset) \\
\addlinespace

LRM-LLaVA (2025) & 
LLaVA-1.5 (with multilingual LLM) & 
machine translated & 
pre-training + instruction tuning & 
10 languages \\
\addlinespace

Chitrarth (2025) & 
LLaVA (with Krutrim 13B LLM) & 
machine translated & 
adapter pre-training + instruction tuning & 
English + 10 Indian languages \\
\addlinespace
\hline
\end{tabular}
\end{table}

A significant amount of work focused on adapting VLMs to low-resource languages and narrowing the performance gap. Table~\ref{tab:vlm_low_resource} summarizes the compared models, architectural choices, the strategy used to gather training data, training method, and covered languages.

LLaVA-NDiNO \citep{musacchio2024llava} represents a family of VLMs trained for the Italian language. The architecture closely follows the LLaVA-NeXT architecture: the authors used CLIP's ViT for the vision encoder, a multi-layer perceptron for the projection module, and LLaMA 3 8B base for the language model. The training dataset represents a mix of Italian native multimodal data and a comprehensive collection of VQA and instructions datasets, machine-translated to Italian. Regarding the training method, the authors followed a four-step procedure. First, they pre-trained the projection module to align visual representations with the LLM’s token space. Second, they trained all modules jointly on Italian image-text pairs. Third, they froze the vision encoder and performed instruction tuning. Finally, they continued instruction tuning on longer, more descriptive samples. The authors reported improved performance on Italian downstream tasks, demonstrating that machine translation can effectively improve the multilingual capabilities of vision-language models.

The Vietnamese language has also been explored through the LaVY \citep{tran2024lavy} model. The authors used a Vietnamese pre-trained LLM and adopted a fine-tuning procedure similar to that of LLaVA-NDiNO. They first pre-trained the projection module to generate tokens compatible with the new LLM, followed by instruction tuning with the vision encoder frozen. In the second stage, they applied LoRA, significantly reducing the computational cost of fine-tuning the LLM. Regarding data preparation, the authors leveraged LLMs at multiple stages: automatically translating English datasets into Vietnamese, refining the translations for fluency and accuracy, and generating image descriptions for locally relevant Vietnamese content. The authors reported state-of-the-art results on Vietnamese downstream tasks.

Language adaptation for Korean was also explored by \citet{shin2024x}, who introduced the bilingual model X-LLaVA, built on top of LLaVA-1.5. Since the original language backbone was trained mostly in English, the authors expanded the vocabulary with Korean tokens to prevent the large linguistic difference from degrading performance. Training was performed in two stages: first, only the MLP projector was aligned, and then both the projector and the LLM were instruction-tuned. The alignment data was machine-translated, while the multilingual instruction-tuning data was generated using GPT-4V. Finally, the authors showed that X-LLaVA achieves significant performance improvements on Korean tasks.

There is also work that aims to improve performance in low-resource languages with multilingual VLMs. LRM-LLaVA \citep{li2025lrm} addressed nine non-English languages from four language families: Sinitic (Chinese), Germanic (German), Romance (Italian, French, Spanish, Romanian), and Slavic (Russian, Czech, Croatian). Each non-English language has an equal share in the training data. The authors introduced, within a LLaVA-1.5-style architecture, a cross-modal regularizer component to better align multilingual textual representations with visual representations. For training, they also employed a suite of multilingual visual question-answering tasks to narrow the multilingual gap. Regarding data, the approach relied on automated machine translation to construct the required multilingual datasets. The authors reported that LRM-LLaVA improves performance on non-English languages while maintaining its English scores, proving the effectiveness of their method.

PALO \citep{maaz2024palo} is another multilingual VLM based on the LLaVA-1.5 architecture. It targets languages with high population coverage across the following families: Germanic (English), Romance (Spanish, French), Slavic (Russian), Sinitic (Chinese), Japonic (Japanese), Indo-Aryan (Hindi, Bengali, Urdu), and Semitic (Arabic). The dataset is created using a semi-automated pipeline: initial LLM-based translation followed by rule-based cleanup of language-specific errors and manual refinement. PALO training followed two steps: projector pretraining, then instruction fine-tuning, where LoRA is applied to the LLM to reduce computational cost. The authors report extensive evaluation showing that multilingual training returns substantial gains for low-resource languages (e.g., Arabic, Hindi, Bengali, Urdu) while sometimes slightly hurting the performance on languages with more resources (e.g., English, French). They attribute this to translation noise in the generated multilingual data and also to the high imbalance in the LLM’s pretraining corpus that favors the high-resource languages, resulting in stronger baselines.

The same language subset of PALO was targeted by Maya \citep{alam2024maya}. The authors translated the original LLaVA dataset into seven languages, covering those in PALO except Bengali and Urdu. However, Maya was also fine-tuned on PALO’s corpus, thus covering all ten PALO languages. In terms of architecture, Maya follows LLaVA-1.5 but swaps the LLM with the multilingual Aya-23 8B \citep{aryabumi2024aya} and the vision encoder with SigLIP \citep{zhai2023sigmoid}. For fine-tuning, the authors experimented with LoRA, but it returned worse results than full fine-tuning. The authors reported strong multilingual performance, surpassing PALO in several languages.

Multilingual VLMs were also proposed for Indian languages. \citet{khan2025chitrarth} introduced Chitrarth, a multilingual VLM targeting English and ten widely spoken languages in India: Assamese, Bengali, Gujarati, Hindi, Kannada, Malayalam, Marathi, Odia, Tamil, and Telugu. The authors followed the LLaVA-1.5 architecture and adopted the standard two-stage training recipe consisting of MLP alignment followed by instruction tuning. Training data were obtained by machine translation of the original LLaVA datasets into each of the ten languages. Experimental results showed that Chitrarth achieved state-of-the-art performance on downstream tasks for low-resource languages, although this comes with a slight impact on English performance.

\subsection{Datasets}

To fine-tune these models and achieve significant gains, task-specific and language-specific datasets are necessary. Because high-quality data is scarce and costly, researchers frequently use synthetic generation and augmentation to create or expand training sets.

\textbf{Machine translation} has been successfully applied by many of the works \citep{musacchio2024llava,tran2024lavy,shin2024x,li2025lrm,maaz2024palo,alam2024maya,khan2025chitrarth}. It introduces relatively lower bias into the model’s training data while requiring substantially less effort than constructing and annotating new datasets. However, the authors usually note several drawbacks for this approach: automatic translations are not perfect and may introduce grammatical or syntactic errors, and translated examples do not preserve the cultural context of the source language. As a consequence, they cannot reflect the linguistic or cultural characteristics of the target language, such as native syntax, idioms, or cultural knowledge.

\textbf{Synthetic data generation} is another widely used set of techniques \citep{li2023llavamed,seyfioglu2023quiltllava,shi-etal-2024-math,lu2024pathchat,gao2025gllava}. The impressive capability of LLMs to understand, reason, and generate new text enables not only the translation of existing datasets from one language to another, but also the transformation of a dataset from one task to another. In this way, a dataset with images and captions that could traditionally be used only for title or description generation can now be converted into an instruction dataset suitable for instruction tuning, offering substantial benefits in terms of cost and speed when creating new resources for low-resource languages. The main drawback of this approach is that it is more prone to introducing bias than simple translation. Depending on the LLM selected to generate the instructions, important details may be missed or misinterpreted. This becomes problematic when the generated data is further used to train the next generation of models because the resulting biases will be propagated and amplified.






%% file: 3method.tex
\subsection{Corpora Development}

\begin{figure}[ht]
    \centering
    \includegraphics[width=0.8\linewidth]{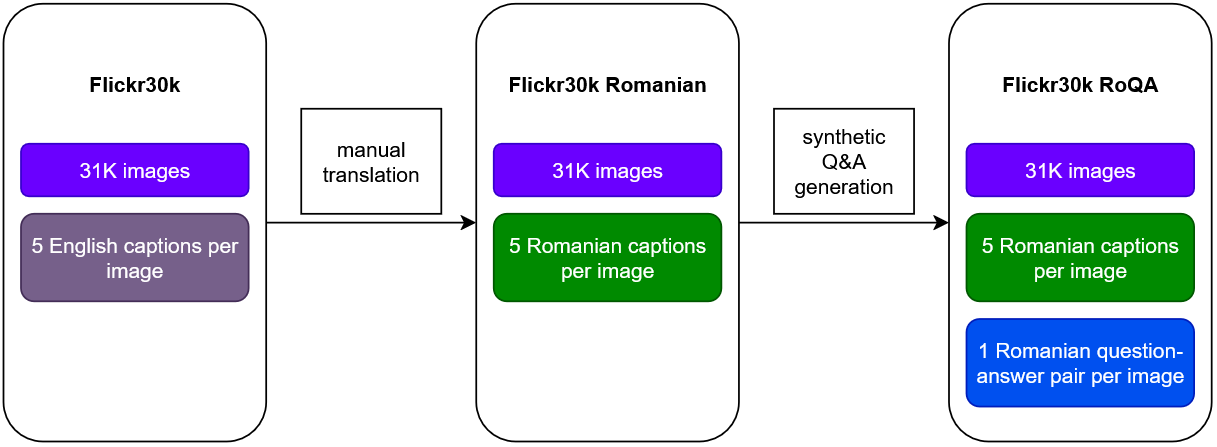}
    \caption{Overview of the dataset construction pipeline.}
    \label{fig:datapipeline}
\end{figure}

We construct the dataset employed in our work, starting with the Flickr30K dataset \citep{young2014image} and following the pipeline illustrated in Figure~\ref{fig:datapipeline}. The original Flickr30K dataset comprises 31,783 images sourced from Flickr\footnote{\url{https://www.flickr.com}, accessed on \accessDate{}}. Each image has five English captions, independently written by humans, that describe its visual content, for a total of 158,915 captions. The images cover a wide range of everyday scenarios, including people engaged in leisure activities or routine tasks, as well as animals and objects. Flickr30k, in the current format, supports several tasks, including image captioning, image-text retrieval, or multimodal representation learning.

\subsubsection{\flickrRO{}}

We create the \flickrRO{} dataset by translating the entire set of English captions from Flickr30K into Romanian. Three native Romanian speakers proficient in English participated in the translation process. Each translator received a subset of approximately 10,600 images and the corresponding 53,000 English captions. The translators’ focus was on preserving meaning and maintaining grammatical correctness. The final dataset follows the structure of the original Flickr30K dataset, comprising 31,783 images each associated with five Romanian captions, for a total of 158,915 sentences. The translated captions are organized in TSV format, with each entry containing the image identifier plus the caption identifier. 

\subsubsection{Extending \flickrRO{} for Visual Question Answering}
\label{sec:extending_flickrro}

To extend the \flickrRO{} dataset for visual question answering, we generate question-answer pairs using only textual descriptions of the images. For each image, we concatenate its five Romanian captions into a single descriptive passage. We refer to the resulting dataset as \flickrROQA{}. 

We selected the largest language model that could fit within the available GPU resources at the time: LLaMA 3.3 70B. The model was hosted using an Ollama\footnote{\url{https://github.com/ollama/ollama}, accessed on \accessDate{}} server running on an A100 GPU with 80 GB of VRAM. The model was prompted to generate four candidate questions per description, focusing on objects, attributes, spatial relations, and actions depicted in the image. One question was then randomly selected from the four candidates to generate answers. In a separate prompt, the model was instructed to provide a concise answer using only the information contained in the description.
Table~\ref{tab:roqa_example} illustrates this process while the prompts are provided in Appendix~\ref{appendix:prompts_used}.

This pipeline produced 31,783 question-answer pairs, matching the number of images in the dataset. The extended dataset supports Romanian-language VQA evaluation and analysis of visually grounded reasoning in Romanian.
We adopt a random train-test split, allocating 80\% of the data for training (25,426 samples) and 20\% for testing (6,357 samples), and use these splits consistently across all experiments reported in this paper. 


\begin{table}[ht]
  \centering
  \caption{Extending \flickrRO{} for Visual Question Answering.}
  \label{tab:roqa_example}
  \begin{tabular}{@{} p{0.55\textwidth}  c @{}}
    \toprule
    \begin{minipage}[t]{\linewidth}
      \vspace{0pt} 

      \textbf{Original English Captions:}\\
    A little gray dog jumps over a bar at an agility test.\\
    A small dog jumping an obstacle in a grassy field.\\
    A little gray dog jumps over a small hurdle.\\
    A little gray dog is jumping over a fence.\\
    A small dog jumping over an obstacle.

      \medskip

      \textbf{Translated Romanian Captions:}\\
    Un mic câine cenușiu sare peste o bară la un test de agilitate.\\
    Un câine mic care sare un obstacol într-un câmp înierbat.\\
    Un mic câine cenușiu sare peste un mic obstacol.\\
    Un câine mic cenușiu sare peste un gard.\\
    Un câine mic care sare peste un obstacol.

      \medskip

      \textbf{Generated Question Candidates:}\\
      Care este culoarea câinelui?\\
      Ce face câinele în imagine?\\
      Peste ce sare câinele?\\
      În ce mediu se află câinele?

      \medskip

      \textbf{Final Question \& Answer:}\\
      Q: Ce face câinele în imagine?\\
      A: Câinele sare peste un obstacol.
    \end{minipage}
    &
    \raisebox{-\totalheight}{%
      \includegraphics[width=0.45\textwidth]{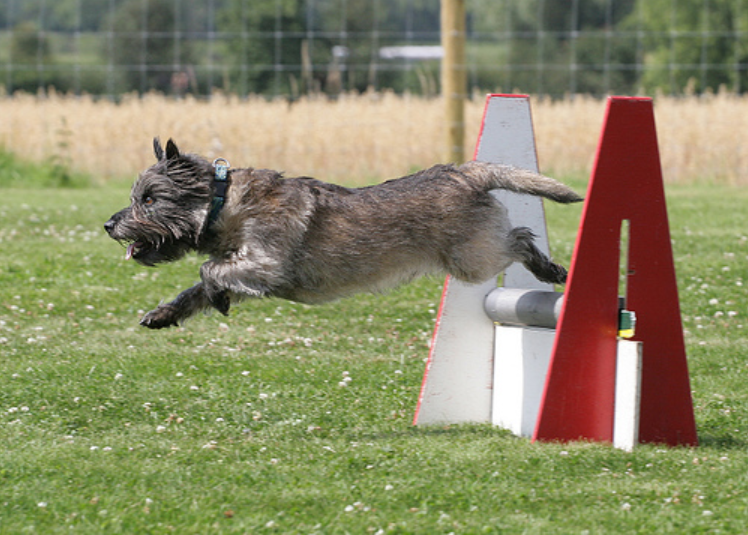}%
    } \\ 
    \bottomrule
  \end{tabular}
\end{table}

\subsubsection{Manual Validation of the \flickrROQA{} Test Set}
\label{sec:manual_validation}

Since the questions and answers in \flickrROQA{} were synthetically generated, we could not correctly assess whether the models would perform well on real-world human questions in Romanian.
To address this, we manually corrected the test split of the \flickrROQA{} dataset.

A single human annotator reviewed the 6,357 question-answer pairs and manually corrected the question and answer fields. This validation focused on ensuring that each question was answerable from the image and that each answer relied exclusively on visible visual information. The annotator corrected for grammatical and content errors, simplified overly specific formulations when possible, and kept edits minimal. In addition, the answers were normalized to full sentences.

\begin{table}[ht]
\centering
\caption{Manual correction statistics for the \flickrROQA{} test set.}
\label{tab:manual_correction_stats}
\begin{tabular}{l r}
\toprule
\textbf{Statistic} & \textbf{Value} \\
\midrule
Test samples & 6,357 \\
Corrected samples & 5,124 \\
Question-only corrections & 262 \\
Answer-only corrections & 1,871 \\
Question-answer corrections & 2,991 \\
Average question edit distance & 13.2 \\
Average answer edit distance & 25.0 \\
\bottomrule
\end{tabular}
\end{table}

Table~\ref{tab:manual_correction_stats} summarizes the manual correction process for the \flickrROQA{} test set. The manual review affected a large portion of the data, indicating that automatic generation required substantial post-processing. Most edits involved the answer field, either alone or together with the question, reflecting the need to improve answer completeness, consistency, and visual grounding. Question-only edits were comparatively less frequent, suggesting that the generated questions were more stable than the answers.

Following prior work that uses CLIPScore to estimate image-text alignment \citep{hessel2021clipscore,basnet2025evaluating}, we compute image-question and image-answer CLIPScores before and after manual correction. Formally, let $I_i$ denote an image from the dataset, $X_i^b\in\{Q_i^{b}, A_i^{b}\}$ represent the generated question-answer text before correction, and $X_i^a\in\{Q_i^{a}, A_i^{a}\}$ are the corresponding human-corrected versions. We thus define the correction gain as:

\begin{equation}
\Delta CLIP_{X_i} = CLIP(I_i, X_i^{a}) - CLIP(I_i, X_i^{b}), 
\quad X_i \in \{Q_i, A_i\}.
\end{equation}

We evaluate the impact of the corrections using the standard OpenAI CLIP model \citep{radford2021learningtransferablevisualmodels} for comparison with prior work. We also use Jina-CLIP v2 \citep{koukounas2025jinaclipv2multilingualmultimodalembeddings}, a multilingual image and text embedding model better suited to non-English text.

\begin{table}[ht]
\centering
\caption{Image and text CLIPScore before and after manual correction.}
\label{tab:clipscore_before_after}
\begin{tabular}{llrrr}
\toprule
\textbf{Model} & \textbf{Pair} & \textbf{Before} & \textbf{After} & \textbf{$\Delta$} \\
\midrule
OpenAI CLIP & Image - Question & 13.016 & 12.976 & -0.040 \\
OpenAI CLIP & Image - Answer   & 13.255 & 13.183 & -0.072 \\
\midrule
Jina-CLIP v2 & Image - Question & 26.807 & 26.583 & -0.223 \\
Jina-CLIP v2 & Image - Answer   & 23.986 & 26.907 & 2.921 \\
\bottomrule
\end{tabular}
\end{table}

Table~\ref{tab:clipscore_before_after} reports the CLIPScore results before and after manual correction. Given that Jina-CLIP v2 appears to better capture changes in Romanian, we focus on these scores and use a paired-samples $t$-test to assess whether the before-and-after CLIPScore differences are statistically significant for the same image-text pairs. For image and question alignment, we observe a small but significant decrease after manual correction, from 26.807 to 26.583 ($\Delta=-0.223$; $t=-4.885$, $p=1.057 \times 10^{-6}$). This reduction is expected, as the manual correction process often removes unimportant visual details from the questions introduced from the original image descriptions. In contrast, image and answer alignment improves substantially after correction, increasing from 23.986 to 26.907 ($\Delta=2.921$; $t=36.286$, $p=3.499 \times 10^{-262}$). This improvement is consistent with the correction objective, which aimed to normalize answers into complete sentences and avoid overly short responses, thereby better aligning the corrected answers with the image content.

\subsubsection{Dataset Statistics}

\begin{figure*}[ht]
    \centering

    \begin{subfigure}{0.49\linewidth}
        \centering
        \includegraphics[
        width=\linewidth,
        trim=0 0 0 0,
        clip
    ]{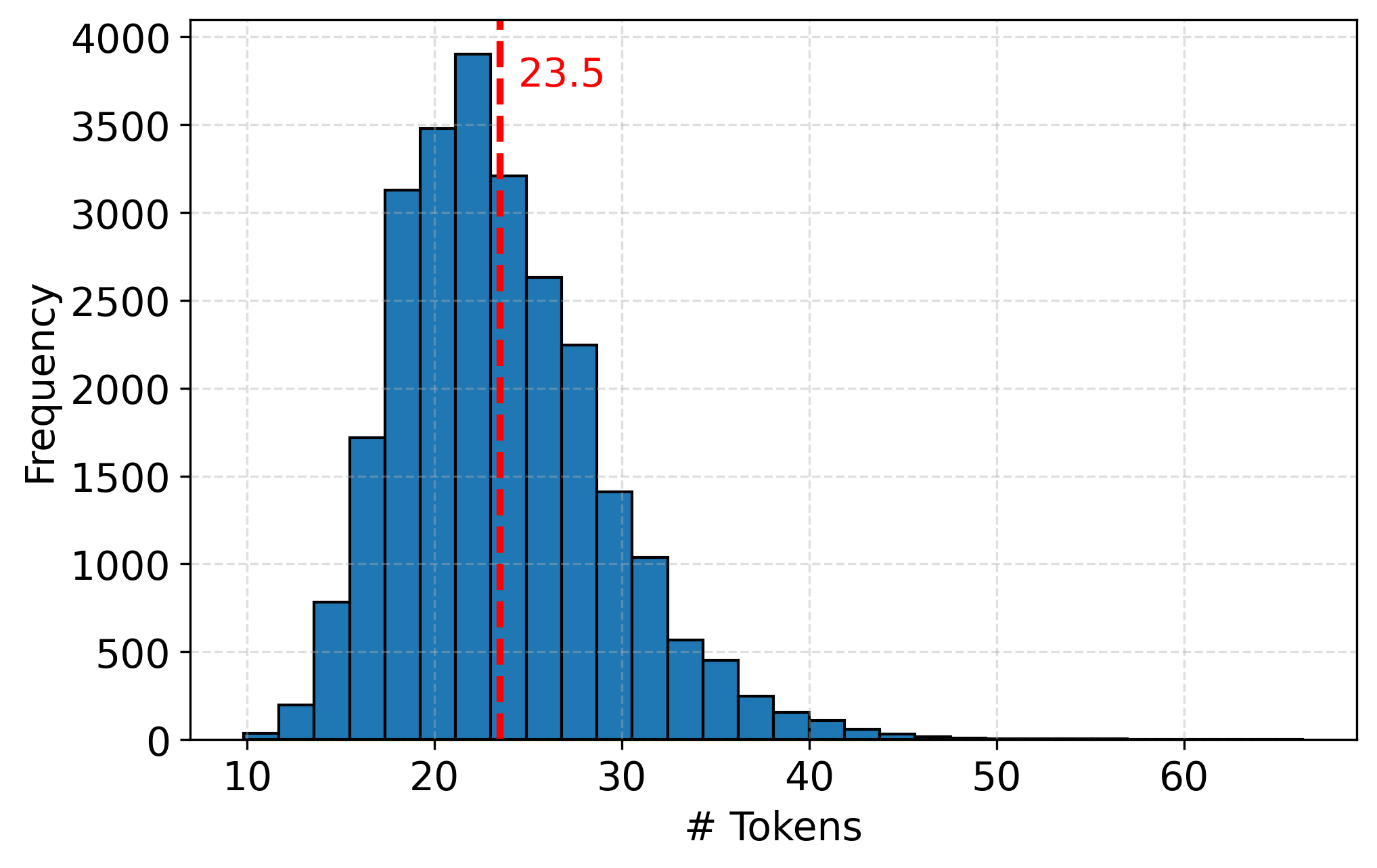}
        \caption{Train captions}
    \end{subfigure}
    \hfill
    \begin{subfigure}{0.49\linewidth}
        \centering
        \includegraphics[
        width=\linewidth,
        trim=0 0 0 0,
        clip
    ]{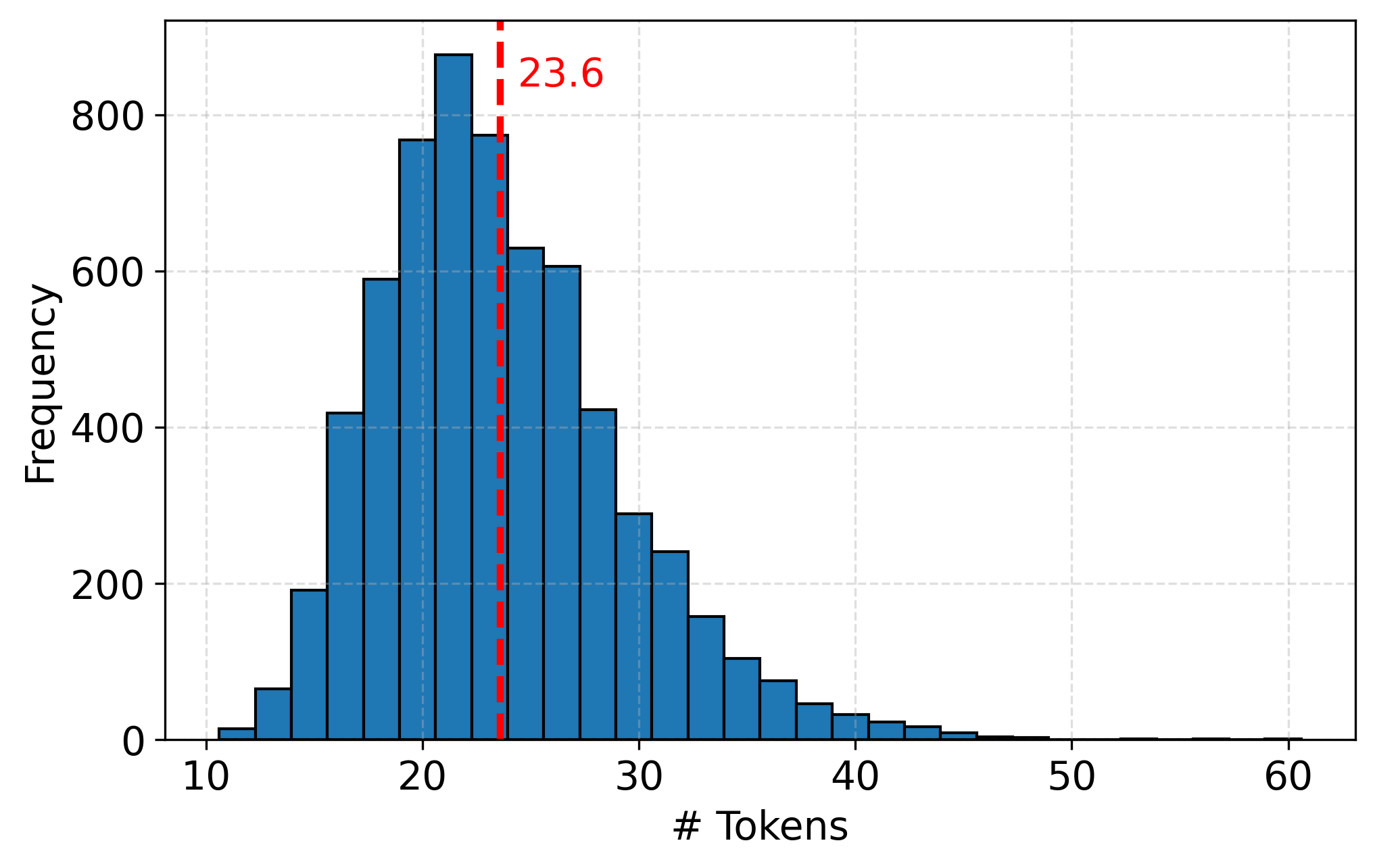}
        \caption{Test captions}
    \end{subfigure}

    \vspace{0.5em}

    \begin{subfigure}{0.49\linewidth}
        \centering
        \includegraphics[
        width=\linewidth,
        trim=0 0 0 0,
        clip
    ]{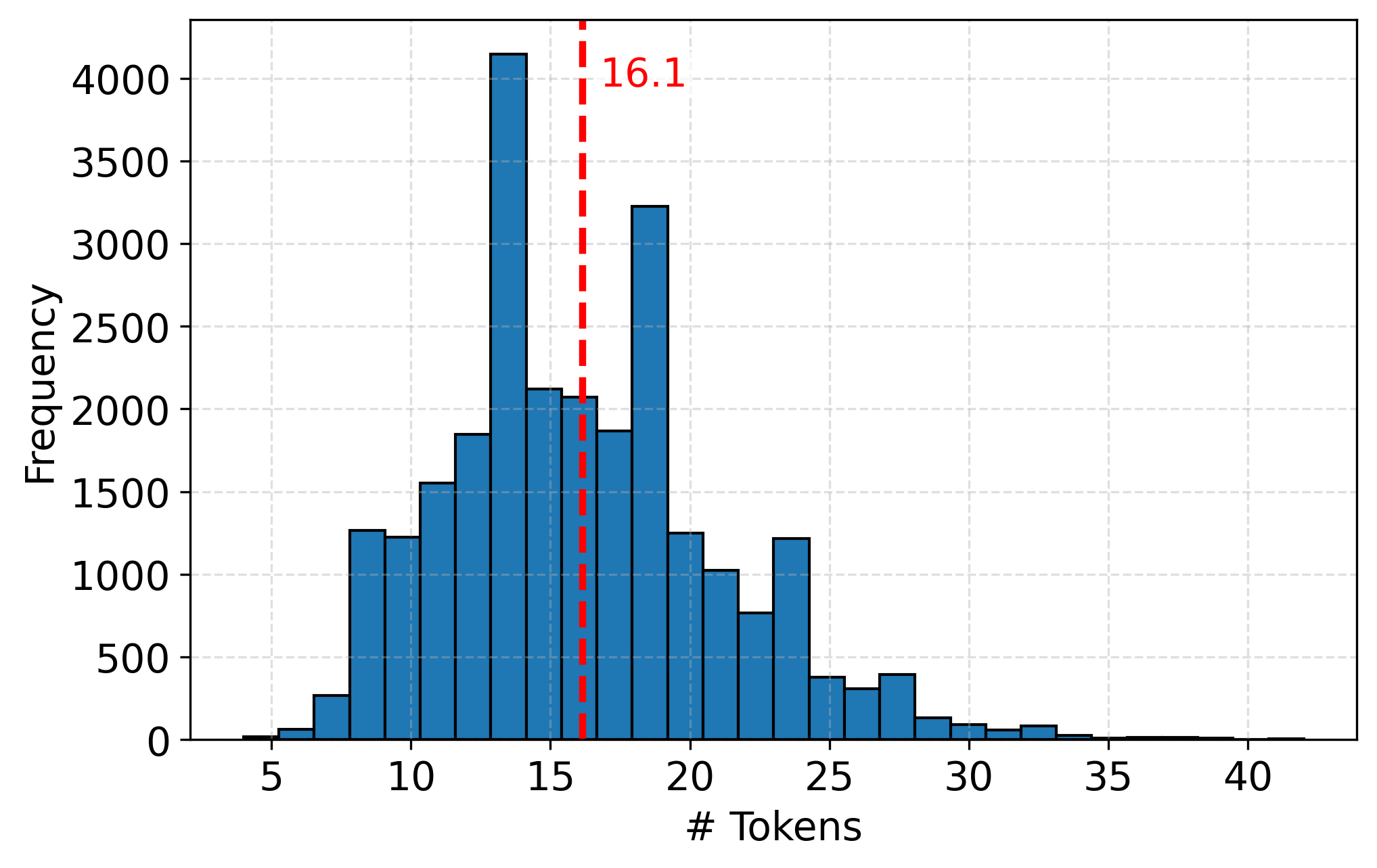}
        \caption{Train questions}
    \end{subfigure}
    \hfill
    \begin{subfigure}{0.49\linewidth}
        \centering
        \includegraphics[
        width=\linewidth,
        trim=0 0 0 0,
        clip
    ]{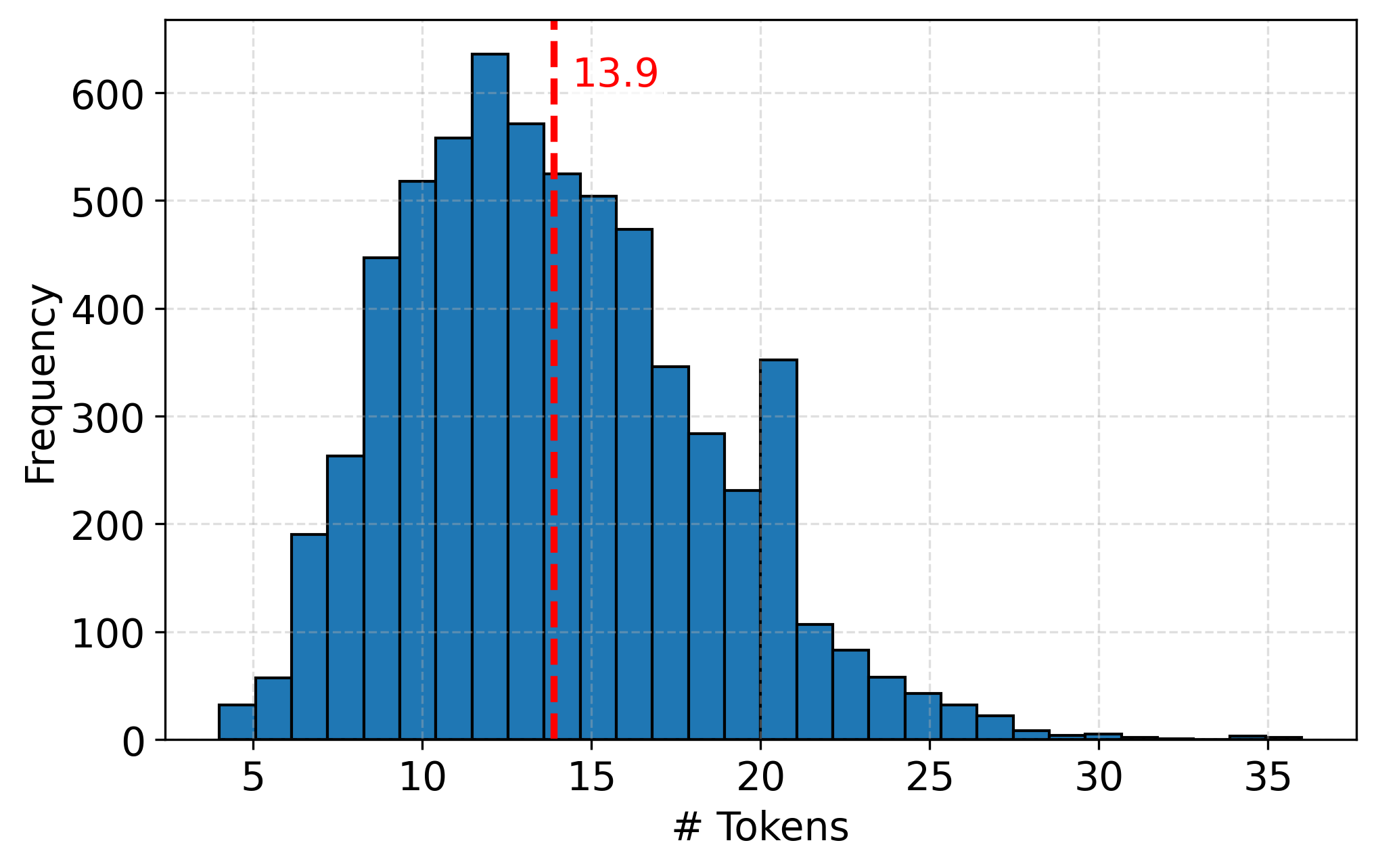}
        \caption{Test questions}
    \end{subfigure}

    \vspace{0.5em}

    \begin{subfigure}{0.49\linewidth}
        \centering
        \includegraphics[
        width=\linewidth,
        trim=0 0 0 0,
        clip
    ]{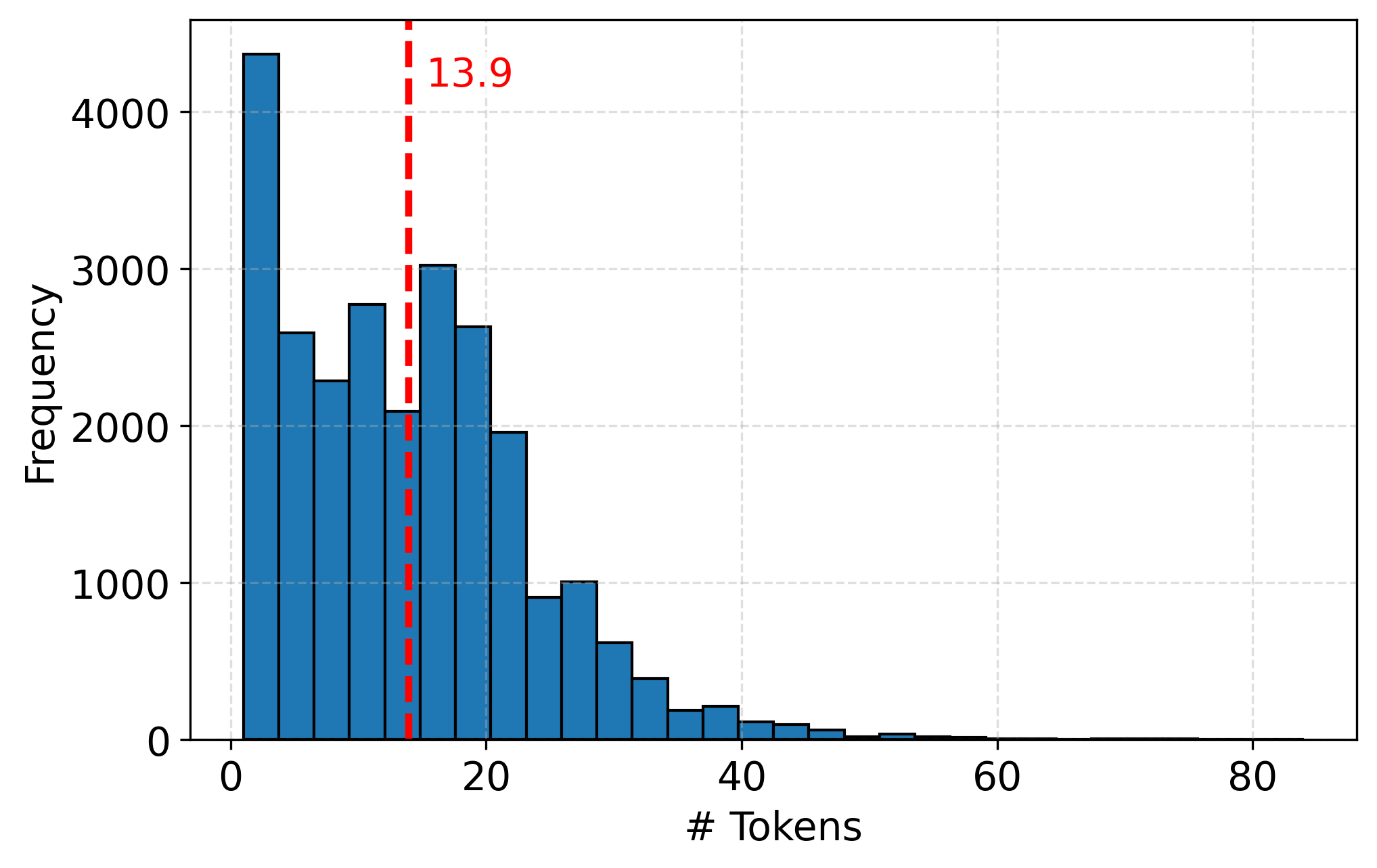}
        \caption{Train answers}
    \end{subfigure}
    \hfill
    \begin{subfigure}{0.49\linewidth}
        \centering
        \includegraphics[
        width=\linewidth,
        trim=0 0 0 0,
        clip
    ]{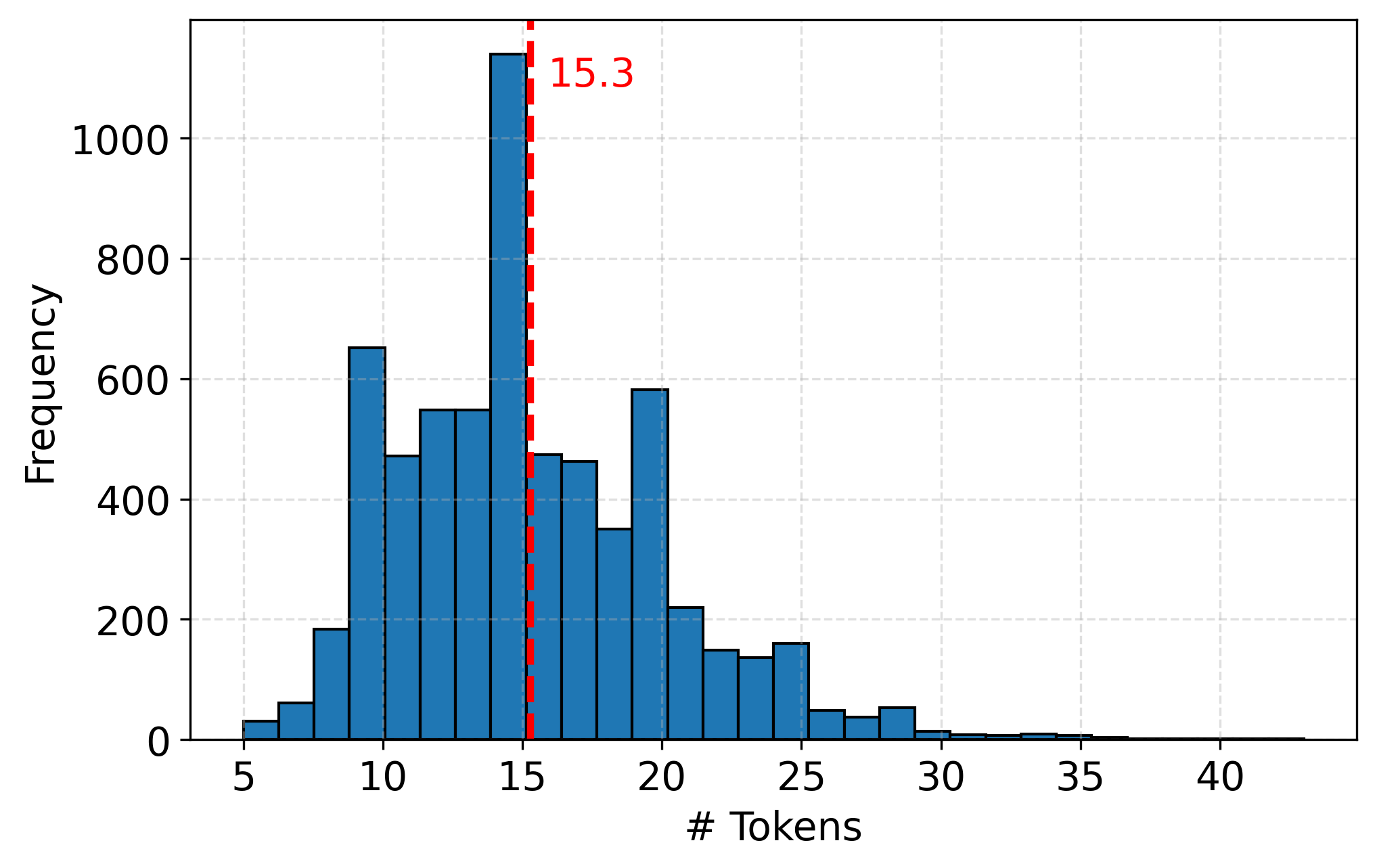}
        \caption{Test answers}
    \end{subfigure}

    \caption{Token count distributions reported for the training and test splits.}
    \label{fig:token_count_distributions}
\end{figure*}

We further analyze the main statistics of the \flickrROQA{} dataset, which also contains the \flickrRO{} dataset. In Figure \ref{fig:token_count_distributions}, we examine the token count distributions for Romanian captions, questions, and answers, reported separately for the training and test splits. Since each image has five captions, we compute the average caption token length per sample. Tokenization is performed using the multilingual cased BERT base model\footnote{\url{https://huggingface.co/google-bert/bert-base-multilingual-cased}, accessed on \accessDate{}} \citep{devlin2019bert}

The captions reveal a unimodal distribution, with an average length of approximately 23-24 tokens and some longer descriptions extending beyond 60 tokens. 
The questions are shorter than captions, with an average of 16.1 tokens in the training split and 13.9 tokens in the test split. The shorter test questions are consistent with the manual correction process, which removed unnecessary details.
The answers have an average length of 13.9 tokens in the training split and 15.3 tokens in the test split. The training answers still contain a visible concentration of very short responses. This spike corresponds to short-form answers, while the remaining reflects full-sentence, more descriptive answers. Because the QA generation prompt did not explicitly state the answer format, the model produced responses in both variants. In contrast, manually corrected test answers are more centered on full-sentence responses, which is expected after the normalization step applied during validation.

\begin{table}[ht]
\centering
\small
\setlength{\tabcolsep}{5pt}
\caption{Top TF-IDF words in Romanian captions.}
\begin{tabular}{l l r @{\hspace{12pt}} l l r}
\toprule
\multicolumn{3}{c}{\textbf{Train}} & \multicolumn{3}{c}{\textbf{Test}} \\
\cmidrule(lr){1-3}\cmidrule(lr){4-6}
\textbf{Word} & \textbf{Translation} & \textbf{TF-IDF} &
\textbf{Word} & \textbf{Translation} & \textbf{TF-IDF} \\
\midrule
bărbat & man     & 0.04080 & bărbat & man     & 0.04257 \\
femeie & woman   & 0.03851 & femeie & woman   & 0.03858 \\
doi    & two     & 0.03030 & doi    & two     & 0.03084 \\
stă    & stands  & 0.02656 & oameni & people  & 0.02834 \\
oameni & people  & 0.02622 & stă    & stands  & 0.02812 \\
om     & man     & 0.02550 & om     & man     & 0.02780 \\
cămașă & shirt   & 0.02458 & cămașă & shirt   & 0.02478 \\
câine  & dog     & 0.02278 & timp   & time    & 0.02342 \\
timp   & time    & 0.02259 & câine  & dog     & 0.02278 \\
lângă  & next to & 0.02125 & lângă  & next to & 0.02154 \\
\bottomrule
\end{tabular}
\label{tab:tfidf_captions}
\end{table}

To provide insight into the content of \flickrROQA{}, Table~\ref{tab:tfidf_captions} highlights the words with the highest TF-IDF scores in the Romanian captions after stopword removal. These terms are words that help distinguish content across samples in the dataset and provide information on the dominant themes present in the corpus. We restrict this analysis to captions, as both questions and answers are generated from the same caption data and therefore largely reflect the same underlying information.

\begin{table}[t]
\centering
\small
\setlength{\tabcolsep}{4pt}
\caption{Top-10 most frequent first words in Romanian questions.}
\begin{tabular}{l l r r @{\hspace{12pt}} l l r r}
\toprule
\multicolumn{4}{c}{\textbf{Train}} & \multicolumn{4}{c}{\textbf{Test}} \\
\cmidrule(lr){1-4}\cmidrule(lr){5-8}
\textbf{Word} & \textbf{Translation} & \textbf{Count} & \textbf{\%} &
\textbf{Word} & \textbf{Translation} & \textbf{Count} & \textbf{\%} \\
\midrule
ce    & what        & 8,222  & 32.34 & ce    & what        & 2,849 & 44.82 \\
care  & which       & 4,694  & 18.46 & unde  & where       & 868   & 13.65 \\
unde  & where       & 2,843  & 11.18 & care  & which       & 447   & 7.03 \\
în    & in          & 2,108  & 8.29  & câte  & how many    & 413   & 6.50 \\
câte  & how many    & 2,099  & 8.26  & în    & in          & 356   & 5.60 \\
cine  & who         & 1,282  & 5.04  & pe    & on          & 334   & 5.25 \\
pe    & on          & 1,192  & 4.69  & cine  & who         & 307   & 4.83 \\
câți  & how many    &   678  & 2.67  & câți  & how many    & 227   & 3.57 \\
cum   & how         &   507  & 1.99  & cum   & how         & 151   & 2.38 \\
este  & is          &   287  & 1.13  & cu    & with        & 67    & 1.05 \\
\bottomrule
\end{tabular}
\label{tab:question_first_word_distribution}
\end{table}

In Table \ref{tab:question_first_word_distribution}, we report the words that appear most frequently at the beginning of the questions in \flickrROQA{}. As expected, most questions start with interrogative terms such as ``ce'' (engl., ``what''), ``care'' (engl., ``which''), and ``unde'' (engl., ``where''). The distribution indicates a good coverage of question types, including spatial, counting, object-focused, and subject-focused questions, as well as a balanced use of different interrogative forms. This suggests that the strategy of generating multiple candidate questions per image and randomly selecting one effectively achieves variability. Also, the first-word distributions across the training and test splits are almost identical, confirming that the data partition preserves the statistical properties of the question set.

\subsection{Baseline Models}

After constructing the Romanian multimodal dataset, we selected three large vision-language models to establish baseline results. We aimed for instruction-tuned architectures that fit our GPU resources and offered strong generalization with limited examples. We excluded very small variants due to their limited performance on visual reasoning tasks. Therefore, the chosen models are \llamaBase{} \citep{grattafiori2024llama3herdmodels}, \qwenBase{} \citep{wang2024qwen2}, and \llavaBase{} \citep{liu2024llavanext}.

\llamaBase{} uses a vision encoder to extract image features, which are then incorporated into the language model through a cross-attention adapter. This design allows the decoder to leverage image representations during generation.

\qwenBase{} uses a Vision Transformer to encode images into visual tokens. It supports dynamic-resolution inputs, producing variable-length visual-token sequences depending on the image size, and uses Multimodal Rotary Position Embedding to combine textual and visual information in a single sequence.

\llavaBase{} combines a CLIP vision encoder \citep{radford2021learningtransferablevisualmodels} with the Mistral-7B language model through a trainable MLP projector. The projector maps image features into the language model's embedding space, enabling the decoder to process visual information alongside the textual prompt.

\subsection{Fine-tuning Setup}

We performed a lightweight, parameter-efficient adaptation of the three baseline models to the \flickrROQA{} corpus by training only a set of Low-Rank Adaptation (LoRA) \citep{hu2022lora} weights.  All original vision parameters remained frozen, while the linear layers of the language model were provided with LoRA modules. This preserves the pretrained visual representations and reduces GPU memory consumption.

We applied supervised fine-tuning to a conversation-style prompt. Each training example is converted into a two-turn chat. The user turn contains the Romanian instruction prompt, followed by the natural-language question and the raw image, while the assistant turn contains the ground-truth answer.

Regarding the LoRA configuration, we set the rank and scaling factor to $r=\alpha=16$, used no dropout, and injected adapters into all linear projections of the language transformer.  This results in a total of $\approx$\,52M trainable parameters (0.48\% of the backbone) for the 11B model, 41M (0.60\%) for LLaVA, and 40M (0.58\%) for Qwen.

Training was performed on a single NVIDIA~A100 80 GB GPU.  Table~\ref{tab:hparams} summarizes the hyperparameters shared across models.

\begin{table}[ht]
  \centering
  \caption{Fine-tuning hyperparameters.}
  \label{tab:hparams}
  \begin{tabular}{@{}ll@{}}
    \toprule
    \textbf{Hyperparameters} & \textbf{Value} \\
    \midrule
    Effective batch size & 16 (2 samples per device $\times$ 8 grad.\ accum.\ steps)\\
    Max steps & 1,500\,($\approx$1 epoch) \\
    Learning rate & $2{\times}10^{-4}$ (linear schedule, 80 warm-up steps)\\
    Optimiser & AdamW 8-bit, weight decay 0.01\\
    Max sequence length & 2048 tokens (text + image)\\
    \bottomrule
  \end{tabular}
\end{table}

Implementation leveraged the FastVisionModel from Unsloth\footnote{\url{https://unsloth.ai/}, accessed on \accessDate{}} 
together with SFTTrainer from TRL\footnote{\url{https://github.com/huggingface/trl}, accessed on \accessDate{}}. One full run on the largest model required roughly 4 hours and 40 minutes, validating the efficiency of the LoRA scheme compared with full-model fine-tuning.


%% file: 4results.tex
\subsection{Evaluation}
\label{sec:evaluation}

We evaluate the fine-tuned VLMs and compare them against the original VLMs on the following tasks: Romanian visual question answering, Romanian image captioning, grammatical error analysis for Romanian, and hallucination analysis under image perturbations. For both Romanian VQA and Romanian image captioning tasks, we evaluate the VLMs using three widely adopted metrics: BLEU, ROUGE-L, and BERTScore.

The three metrics complement each other by offering distinct perspectives on VLMs' performance. BLEU focuses on precision and rewards shared n-grams between the generated and reference texts. ROUGE-L captures how well the generated text covers the content of the reference by measuring the longest common subsequence between the VLM output and the reference. Finally, BERTScore assesses semantic similarity, indicating how closely the meanings of the two texts align. All scores are reported as percentages except BLEU, which is shown in raw corpus units.

\textbf{BLEU.}
Bilingual Evaluation Understudy (BLEU) is a metric originally proposed by \citet{papineni2002bleu} to assess the quality of automatic translations. Specifically, BLEU (Equation~\ref{eq:bleu}) computes the geometric mean of the first 
$N=4$ n-gram precisions between the candidate text $C$ and the reference text $R$, which is then scaled by a brevity penalty (Equation~\ref{eq:brevity-penalty}). Formally,

\begin{equation}
\text{BLEU} =
\text{BP} \cdot 
\exp\left( \sum_{n=1}^{N} \frac{1}{N} \log P_n \right)
\label{eq:bleu}
\end{equation}

\begin{equation}
\text{BP} = 
\left\{
\begin{array}{ll}
1, & \text{if } |C| > |R| \\[6pt]
\exp\left( 1 - \frac{|R|}{|C|} \right), & \text{if } |C| \le |R|
\end{array},
\right.
\label{eq:brevity-penalty}
\end{equation}

where $|C|$ and $|R|$ denote the length of the candidate text $C$ and reference text $R$, respectively.

\textbf{ROUGE.}
Recall-Oriented Understudy for Gisting Evaluation (ROUGE) was introduced by \citet{chin2004rouge} to address the automatic evaluation of summaries. In this work, we use ROUGE-L, a variant based on the longest common subsequence (LCS). More specifically, given a reference text 
$R$ and a candidate text $C$, the longest common subsequence 
$LCS(R,C)$ is first computed. Based on the $LCS$, we derive the precision and recall (Equation~\ref{eq:rouge-rec}), which are then combined into the ROUGE-L F-score (Equation~\ref{eq:rouge-f1}) using 
$\beta=1$. This corresponds to the harmonic mean of precision and recall:

\begin{equation}
P_{\mathrm{LCS}} = \frac{\mathrm{LCS}(R, C)}{|C|},
\quad
R_{\mathrm{LCS}} = \frac{\mathrm{LCS}(R, C)}{|R|},
\label{eq:rouge-rec}
\end{equation}

\begin{equation}
\mathrm{ROUGE\text{-}L} =
2 \cdot \frac{P_{\mathrm{LCS}} \cdot R_{\mathrm{LCS}}}
{R_{\mathrm{LCS}} + P_{\mathrm{LCS}}}
\label{eq:rouge-f1}
\end{equation}

\textbf{BERTScore.} 
BERTScore \citep{zhang2019bertscore} was proposed as an automatic method for evaluating text generation tasks. It computes the similarity between a reference text $R$ and a candidate text $C$ by measuring the similarity of their token-level contextual embeddings $r \in R$ and $c \in C$, respectively. Firstly, token embeddings are generated with BERT \citep{devlin2019bert}, and then precision and recall (Equation~\ref{eq:bert-rec}) are computed based on the maximum cosine similarity between token-embedding pairs. Finally, the F\textsubscript{1} score is computed (Equation~\ref{eq:bert-f1}):

\begin{equation}
P_{\text{BERT}} = \frac{1}{|C|} \sum_{c \in C} \max_{r \in R} r^\top c,
\quad
R_{\text{BERT}} = \frac{1}{|R|} \sum_{r \in R} \max_{c \in C} r^\top c,
\label{eq:bert-rec}
\end{equation}

\begin{equation}
BERTScore F_{1} =
2 \cdot \frac{P_{\text{BERT}} \cdot R_{\text{BERT}}}
{P_{\text{BERT}} + R_{\text{BERT}}}.
\label{eq:bert-f1}
\end{equation}

\textbf{Grammatical Correctness.}
For the grammatical error analysis, we use standard metrics commonly used in other grammatical error-correction tasks: the number of substituted (\(S\)), inserted (\(I\)), and deleted (\(D\)) words. These errors are combined into the widely used Word Error Rate (WER), defined as
\begin{equation}
\text{WER} = \frac{S + I + D}{N},
\end{equation}
where \(N\) denotes the total number of words in the reference text.

\textbf{Hallucination Metrics.}
In order to assess model robustness under perturbations, we employ object-level hallucination
metrics. Specifically, we use \emph{Caption Hallucination Assessment with Image Relevance}
(CHAIR), originally introduced by \citet{rohrbach2019objecthallucinationimagecaptioning},
and its binary derivative \emph{Hal}, as defined by
\citet{ding2024hallupievaluatinghallucinationmultimodal}.

Given a candidate response $C$ and a reference text $R$, we first extract the sets of objects appearing in each text, denoted by $C_{\text{obj}}$ and $R_{\text{obj}}$,
respectively. The CHAIR score, defined in Equation~\ref{eq:chair}, measures the proportion of
objects mentioned in the candidate that are not supported by the reference, meaning the object hallucination rate:

\begin{equation}
\text{CHAIR} = 1 - \frac{|C_{\text{obj}} \cap R_{\text{obj}}|}{|C_{\text{obj}}|}
\label{eq:chair}
\end{equation}

The Hal metric (Equation~\ref{eq:hal}) is a binary hallucination indicator derived from CHAIR.
A candidate response $C$ is considered hallucinated if its CHAIR score is greater than zero:

\begin{equation}
\text{Hal} =
\begin{cases}
1, & \text{if } \text{CHAIR} > 0, \\
0, & \text{otherwise}.
\end{cases}
\label{eq:hal}
\end{equation}

\subsection{Romanian Visual Question Answering}
\label{sec:romanian_vqa}

The three fine-tuned models, together with their base models, were evaluated for Visual Question Answering on the
\flickrROQA{} dataset. For
each image, the model-generated answer was compared against the reference answer using
BERTScore (precision, recall, and F\textsubscript{1}), ROUGE-L (precision, recall, and F\textsubscript{1}), and BLEU. Please note that automatic metrics are computed against a single reference answer, which may not capture all equivalently correct phrasings, so we use BERTScore to help address this issue.

\begin{table}[ht]
  \centering
  \caption{Results for Romanian Visual Question Answering.}
  \label{tab:results_vqa}
  \begin{tabular}{@{}lccccccccc@{}}
    \toprule
    \multirow{2}{*}{\textbf{Model}} &
      \multicolumn{3}{c}{\textbf{BERTScore (\%)}} &
      \multicolumn{3}{c}{\textbf{ROUGE-L (\%)}} &
      \multicolumn{1}{c}{\textbf{BLEU}} \\
    \cmidrule(lr){2-4} \cmidrule(lr){5-7} 
    & P     & R     & F$_{\text{1}}$    & P      & R      & F$_{\text{1}}$ &  \\
    \midrule
   \llamaBase{}
      & 51.39 & 63.07 & 56.37
      &  8.91 & 32.14 & 11.58 & 0.90 \\
    \llamaRO{}
      & 71.13 & 70.20 & 70.32
      & 44.37 & 40.93 & 38.70 & 21.14 \\
    \llavaBase{}
      & 64.80 & 76.35 & 69.96
      & 32.16 & \textbf{61.55} & 38.30 & 8.25 \\
    \llavaRO{}
      & 66.34 & 78.06 & 71.43
      & 54.07 & 56.73 & 52.82 & 26.05 \\
    \qwenBase{}
      & 71.69 & \textbf{79.44} & 75.09
      & 40.57 & 58.69 & 45.14 & 14.30 \\
    \qwenRO{}
      & \textbf{79.09} & 76.38 & \textbf{77.38}
      & \textbf{62.30} & 53.85 & \textbf{53.54} & \textbf{34.79} \\
    \bottomrule
  \end{tabular}

\end{table}

The results reported in Table \ref{tab:results_vqa} indicate that all three models benefit from parameter-efficient adaptation in terms of BERTScore F\textsubscript{1}, ROUGE-L F\textsubscript{1}, and BLEU, although the magnitude of the gains differs by architecture and evaluation metric. Among the fine-tuned variants, \qwenRO{} attains the highest overall performance, registering a BERTScore F\textsubscript{1} of 77.38\%, a ROUGE-L F\textsubscript{1} of 53.54\%, and a BLEU score of 34.79. Relative to its base model, these values correspond to absolute improvements of 2.29 percentage points, 8.40 percentage points, and 20.49, respectively, suggesting that Qwen's multilingual pre-training can be further strengthened by domain-specific Romanian supervision. Because BERTScore emphasizes semantic correspondence rather than exact n-gram overlap, the simultaneous growth in ROUGE-L and BLEU suggests that the adapter not only preserves meaning but also increases lexical alignment with the gold answers.

\llamaRO{} shows a similar advance, with its BERTScore~F\textsubscript{1} rising from 56.37\% to 70.32\% and its BLEU score increasing from 0.90 to 21.14. The substantial gains in both BLEU and ROUGE-L point to improved n-gram precision and better overlap with longer sequences in the reference answers. Given that only 0.48\% of the backbone parameters were updated, these gains highlight the effectiveness of the LoRA strategy for adapting a high-capacity vision-language transformer to a low-resource target language.

Fine-tuning \llavaBase{} yields consistent improvements across all evaluation metrics. The adapter increases the BLEU score from 8.25 to 26.05, while BERTScore~F\textsubscript{1} rises by 1.47 percentage points, from 69.96\% to 71.43\%, and ROUGE-L~F\textsubscript{1} improves substantially from 38.30\% to 52.82\%. 


Across backbones, BERTScore recall consistently exceeds precision in the base models, suggesting that their answers achieve broader semantic coverage but may also include content not present in the reference answers. Fine-tuning reverses this relationship for LLaMA and Qwen, with precision slightly exceeding recall, indicating a shift towards more selective and reference-aligned outputs. For LLaVA, the recall-precision gap remains nearly unchanged, increasing only marginally after fine-tuning. ROUGE-L values, which capture longest common subsequences, remain lower than BERTScore values in all settings, underlining the greater difficulty of matching the exact lexical choices and word order of the Romanian reference answers.

Model size and pre-training scope also affect baseline scores. Despite having fewer parameters, \qwenBase{} outperforms the larger \llamaBase{} before adaptation across all reported metrics, likely because it was trained on a multilingual corpus that covers roughly thirty languages beyond English and Chinese. The technical report introducing Qwen 2 \citep{wang2024qwen2} states that the pre-training data were assembled ``to foster multilingual proficiency''.
After adaptation, LLaMA narrows the gap but does not surpass \qwenRO{}, which retains a clear advantage across all metrics. The results, therefore, suggest that a modestly sized model with targeted multilingual pre-training can match or exceed a larger monolingual model once both are exposed to task-specific supervision.

Taken together, the findings demonstrate that lightweight LoRA adapters can substantially enhance Romanian VQA performance when mounted on multilingual or English-centric vision-language models. The success of \qwenBase{} underscores the importance of initial language coverage.

\subsection{Romanian Image Captioning}

The six models evaluated for Visual Question Answering were further assessed on the Romanian Flickr30K captioning test split without any additional task-specific fine-tuning.  
The results are presented in Table~\ref{tab:results_captioning}.

\begin{table}[ht]
  \centering
  \caption{Results for Romanian Image Captioning.}
  \label{tab:results_captioning}
  \begin{tabular}{@{}lccccccc@{}}
    \toprule
    \multirow{2}{*}{\textbf{Model}} &
      \multicolumn{3}{c}{\textbf{BERTScore (\%)}} &
      \multicolumn{3}{c}{\textbf{ROUGE-L (\%)}} &
      \multicolumn{1}{c}{\textbf{BLEU}} \\
    \cmidrule(lr){2-4} \cmidrule(lr){5-7}
    & P     & R     & F$_{\text{1}}$    & P      & R      & F$_{\text{1}}$ &  \\
    \midrule
    \llamaBase{}
      & 52.70 & 64.12 & 57.22
      &  9.32 & 33.65 & 14.00 & 1.53 \\
    \llamaRO{}
      & 65.49 & 69.57 & 66.69
      & 27.35 & 33.73 & 28.60 & 6.99 \\
    \llavaBase{}
      & 55.05 & 67.47 & 59.90
      & 10.86 & 40.65 & 16.70 & 2.62 \\
    \llavaRO{}
      & 65.27 & 73.82 & 68.54
      & \textbf{42.16} & 44.36 & \textbf{41.45} & \textbf{15.38} \\
    \qwenBase{}
      & 64.50 & 72.64 & 67.50
      & 23.00 & 39.92 & 28.07 & 7.18 \\
    \qwenRO{}
      & \textbf{69.92} & \textbf{76.14} & \textbf{71.95}
      & 33.18 & \textbf{47.84} & 37.74 & 14.94 \\
    \bottomrule
  \end{tabular}
\end{table}

Although the adapters were trained exclusively on the Romanian VQA corpus, all three architectures transfer the acquired knowledge to caption generation. The LLaMA adapter yields an absolute increase of 5.46 in BLEU, while the Qwen2 adapter improves BLEU by 7.76. The largest improvement is obtained by the LLaVA adapter, whose BLEU score increases by 12.76.
Corresponding gains in BERTScore and ROUGE-L indicate that the captions are not only lexically closer to the references but also semantically more faithful. These findings suggest that parameter-efficient Romanian instruction tuning encourages broader language-vision alignment beyond the task on which the adapters were trained.

The RoVQA adapter consistently benefits \qwenBase{}, which achieves the highest BERTScore precision, recall, and F\textsubscript{1}, as well as the highest ROUGE-L recall, despite having fewer parameters than \llamaBase{}. However, \llavaRO{} shows the strongest exact lexical overlap with the reference captions, achieving the highest ROUGE-L precision and F\textsubscript{1}, as well as the highest BLEU score.

Across all models, BERTScore recall exceeds precision, suggesting that the generated captions achieve broader semantic coverage than semantic precision. Fine-tuning narrows this gap for all tested architectures, indicating a more balanced relationship between semantic precision and recall in the adapted models.
ROUGE-L remains markedly lower than BERTScore,
reflecting the challenge of matching word order and inflection in Romanian.

The experiments demonstrate that a small amount of task-specific Romanian supervision provides measurable benefits on a distinct multimodal task. 

\section{Discussion}
\label{sec:discussions}

\subsection{Synthetic Data and Parameter-Efficient Fine-Tuning}

To answer \textbf{RQ1}, we show in \S\ref{sec:extending_flickrro} that LLMs can be effectively used to generate useful training data for a low-resource language. Our synthetic visual question-answering generation pipeline produces the Romanian VQA corpus by leveraging LLaMA 3.3 70B as a teacher model to bootstrap Romanian multimodal capabilities into smaller VLMs. This is further emphasized in \S\ref{sec:manual_validation} in which human validation showed that fewer manual edits were required for the questions, while the answers, being required to be complete and correct, required more considerable manual work. This observation is also validated by the CLIP scores before and after the manual corrections as illustrated in Table~\ref{tab:clipscore_before_after}.

As suggested in \S\ref{sec:romanian_vqa}, the improvements observed in VQA, together with the transfer gains in image captioning, suggest that LLM-based supervision can help bootstrap language-specific capabilities when data are scarce.

Experimentally, to address \textbf{RQ2}, instruction tuning on our Romanian corpus yields consistent, substantial improvements across all three model families. On visual question answering, the adapted \qwenBase{} model achieves the strongest overall performance. \llamaBase{} shows the largest improvement in BERTScore~F\textsubscript{1}, demonstrating that a modest amount of language-specific supervision can substantially enhance multimodal understanding.

Taken together, our findings indicate that PEFT can substantially improve the low-resource language capabilities of vision-language models. LoRA adaptation shows clear gains across all tested architectures, improving performance on Romanian visual tasks, grammatical fluency, and hallucination behavior.

\subsection{Grammatical Error Analysis of Romanian VQA Answers}

Beyond delivering semantically correct content, a VQA system deployed in real-world applications must formulate answers that are grammatically sound. Consequently, the aim of this section is to quantify the extent to which Romanian-specific instruction tuning mitigates grammatical errors.

Because human-edited references are not available for the Romanian
VQA corpus, we adopt an automatic LLM-as-a-Judge protocol \citep{zheng2023judge}.
Each answer produced by the two LLaMA variants is fed to a stronger, instruction-tuned model (LLaMA 3.3 70B), which returns a grammatically corrected version.
We then align the original and corrected texts with the Jiwer\footnote{\url{https://jitsi.github.io/jiwer}, accessed on \accessDate{}} toolkit and compute the number of substitutions, deletions,
and insertions, together with the overall word-error rate (WER).
Table~\ref{tab:grammar_errors} reports the average counts per sentence as well as WER, aggregated over the entire test set. Lower values indicate better grammatical correctness.

\begin{table}[ht]
  \centering
  \caption{Average grammatical error statistics.}
  \label{tab:grammar_errors}
  \begin{tabular}{@{}lcccc@{}}
    \toprule
    \textbf{Model} & \textbf{Substitutions} & \textbf{Insertions} &
    \textbf{Deletions} & \textbf{WER (\%)} \\
    \midrule
    \llamaBase{}               & 5.57 & 0.85 & 1.04 & 19.72 \\
    \llamaRO{}         & 0.73 & 0.12 & 0.06 & 15.96 \\
    \llavaBase{}               & 6.48 & 1.03 & 1.27 & 21.84 \\ 
    \llavaRO{}         & 0.85 & 0.35 & 0.15 & 21.26 \\ 
    \qwenBase{}                 & 4.42 & 0.64 & 0.78 & 17.35 \\ 
    \qwenRO{}           & \textbf{0.68} & \textbf{0.11} & \textbf{0.05} & \textbf{14.12} \\ 
    \bottomrule
  \end{tabular}
\end{table}

The results in Table~\ref{tab:grammar_errors} show that all three models exhibit fewer grammatical errors in Romanian after visual question answering fine-tuning. The high substitution rates before fine-tuning indicate that many words are grammatically incorrect, likely due to missing definite articles or diacritic loss. LLaMA shows the largest improvement in word error rate, followed by Qwen, while LLaVA shows substantial decreases in edit operations but only a marginal improvement in WER.

Nevertheless, two caveats must be acknowledged. First, WER penalizes all edits equally and therefore fails to distinguish between genuine grammar errors and lexical paraphrases introduced by the judge model. Second, relying on a single LLM as a reference risks propagating its own stylistic preferences.

Overall, for \textbf{RQ3} we find that even a narrow task-specific adapter can confer measurable improvements in linguistic form, implying that content accuracy and grammatical fluency are not mutually exclusive goals.

\subsection{VQA Hallucinations and Robustness to Perturbations}

In this section, we assess the effects of Romanian fine-tuning on hallucinations in visual question answering. Furthermore, inspired by \citet{ding2024hallupievaluatinghallucinationmultimodal}, we evaluate how robust our models are to hallucinations under selected input perturbations. All experiments in this section are conducted on the test split of the Romanian Flickr30K VQA dataset.

To measure hallucinations, we apply the CHAIR metric and its derivative, Hal, as described in \S\ref{sec:evaluation}. CHAIR is computed based on the set of objects extracted from generated answers and reference texts. Since manually annotating the object sets is impractical, we use an automated object extraction procedure. Specifically, we use the spaCy\footnote{\url{https://spacy.io/}, accessed on \accessDate{}} Romanian part-of-speech tagger (i.e., \texttt{ro\_core\_news\_sm}) and keep only tokens classified as common or proper nouns. We note that this introduces a limitation because inaccuracies in POS tagging may affect the resulting hallucination metrics. However, the approach is computationally efficient, and manual inspection of a subset of samples indicates that the extracted object sets are generally reasonable.

After computing CHAIR and Hal for the original Romanian VQA setting, we apply perturbations to analyze how parameter-efficient fine-tuning influences model robustness. Following Hallu-PI \citep{ding2024hallupievaluatinghallucinationmultimodal}, we select image concatenation and cropping perturbations because these target the alignment between vision and language. Other perturbations, such as blur or noise, mainly affect the visual encoder and are therefore less relevant for this paper.

For the concatenation perturbation, we group the samples into sets of four images. Each image is resized to $256\times256$, and the four images are combined into a single concatenated image $512\times512$, with the anchor image placed in the upper left corner. The Romanian question is modified by prefixing a location cue, \emph{``În imaginea din stânga sus,''} (``In the image from the top-left corner,''), while the ground-truth answer is kept unchanged. The three extra images are removed from the dataset as standalone samples to keep the evaluation samples independent. The anchor image is always placed in the top-left position to ensure consistent difficulty across samples, so that observed differences can be attributed solely to the perturbation.

For the cropping perturbation, we remove a $10\%$ margin from each side of the image, resulting in a centrally cropped image having $80\%$ of the original width and height. The question prompt is updated by adding the prefix \emph{``În imaginea decupată,''} (``In the cropped image,''), while the answer remains unchanged. This perturbation tests whether models hallucinate when partial visual information is removed.

\begin{table}[ht]
  \centering
  \caption{VQA Hallucination results under perturbations.}
  \label{tab:hallucination_perturb_deltas}
  \begin{tabular}{@{}lcccccc@{}}
    \toprule
    \textbf{Model} &
    \multicolumn{2}{c}{\textbf{Original}} &
    \multicolumn{2}{c}{\textbf{Concatenation ($\Delta$)}} &
    \multicolumn{2}{c}{\textbf{Cropping ($\Delta$)}} \\
    \cmidrule(lr){2-3}\cmidrule(lr){4-5}\cmidrule(lr){6-7}
    & \textbf{Hal (\%)} & \textbf{CHAIR}
    & \textbf{Hal (pp)} & \textbf{CHAIR}
    & \textbf{Hal (pp)} & \textbf{CHAIR} \\
    \midrule
    \llamaBase{}
      & 97.61 & 0.83 & +0.57 & +0.00 & +0.38 & +0.00 \\
    \llamaRO{}
      & 69.92 & 0.46 & -1.64 & +0.03 & -1.01 & +0.01 \\
    \llavaBase{}
      & 93.08 & 0.89 & +6.61 & -0.20 & +4.66 & +0.06 \\
    \llavaRO{}
      & 91.94 & 0.53 & +1.51 & +0.05 & +2.01 & +0.04 \\
    \qwenBase{}
      & 90.25 & 0.54 & +8.18 & +0.08 & +6.99 & +0.01 \\
    \qwenRO{}
      & 56.58 & 0.32 & +6.36 & +0.06 & +3.52 & +0.02 \\

    \bottomrule
  \end{tabular}
\end{table}

To answer \textbf{RQ4}, the results reported in Table~\ref{tab:hallucination_perturb_deltas} show that Romanian fine-tuning reduces hallucinations on the downstream Romanian VQA task across all three model families. The improvements are particularly pronounced for \llamaBase{} and \qwenBase{}, whereas LLaVA-v1.6 exhibits a modest reduction in Hal but a substantial improvement in CHAIR. Overall, \textbf{\qwenRO{}} achieves the lowest hallucination scores. These findings suggest that fine-tuning can improve robustness to perturbations. However, the extent of this improvement depends on the model architecture.

Under perturbations, we note that the models generally show degraded performance relative to the original setting, with image concatenation posing a greater challenge than cropping. This suggests that multiple visual contexts are more disruptive for VLMs than partial removal of visual information, consistent with the observations of \citet{ding2024hallupievaluatinghallucinationmultimodal}.

Interestingly, the fine-tuned \textbf{\llamaRO{}} model shows a decrease in Hal under both perturbations, despite a small increase in CHAIR. This may indicate that the model hallucinates in fewer answers under perturbation while adding slightly more incorrect objects to the answers that still hallucinate.

\subsection{Limitations}

Despite these promising results, our study has several limitations. First, the synthetic QA component relies on a LLaMA 3.3 generator (primarily English-centric) conditioned only on text, which may omit visually grounded reasoning patterns specific to Romanian usage. Second, our fine‐tuning experiments are confined to three publicly available MLLMs; results may differ for larger architectures.  Third, while our human‐verified translations ensure high surface quality, the scope of linguistic phenomena covered remains limited to everyday photographic scenes.  Scenarios involving complex visual abstractions or domain‐specific terminology are not addressed.

%% file: 5conclusion.tex
In this work, we have presented a systematic study of parameter‐efficient multimodal instruction tuning for Romanian.  By constructing \textbf{\flickrRO{}}, a human‐verified translation of the Flickr30K captions corpus, and extending it with synthetic question-answer pairs to form \textbf{\flickrROQA{}}, we supply a bilingual dataset of 158,915 captions and 31,783 visual QA examples. The VQA test set was additionally validated by a human annotator, who revised some questions and answers where necessary.  Notably, all synthetic QA generation and adapter tuning was carried out exclusively with open‐source models: LLaMA 3.3 70B  served as a ``teacher'' to bootstrap Romanian capabilities into smaller MLLMs. Our approach leveraged LoRA layers inserted only into the language branches of three representative open‐source MLLMs: \llamaBase{}, \qwenBase{}, and \llavaBase{}.  This design yields a lightweight fine-tuning pipeline requiring under five hours on a single A100 GPU and updating at most 0.6 \% of model parameters.

Our findings demonstrate that a carefully curated Romanian vision-language corpus, in conjunction with parameter‐efficient adapter tuning, can substantially mitigate the performance gap for underrepresented languages. 
The grammatical error analysis also shows lower error rates after fine-tuning, with the largest WER reduction observed for \llamaBase{}. In addition, the adapted models generally hallucinate less in the original VQA setting. Furthermore, without additional caption-specific training, the same LoRA adapters generalize to Romanian image captioning, yielding BLEU improvements of up to 12.76 points.
We release all data and adapter weights to facilitate further progress toward truly multilingual and equitable vision-language technologies.  

Future research directions include domain adaptation. We plan to extend the corpus to specialized domains (e.g., medical imaging, remote sensing) by collecting or translating domain‐specific captions and QA pairs, thereby evaluating the generality of LoRA-based tuning.

%% file: appendix.tex
\section{Prompts Used}
\label{appendix:prompts_used}

In Table \ref{tab:prompt_templates}, we present the Romanian prompts used in this paper and their English translations. The question generation prompt is used to produce four candidate questions for each image during the dataset augmentation phase. The \texttt{<image description>} placeholder is replaced with the concatenated image captions. The answer generation prompt is then used to generate answers for randomly selected questions, where the \texttt{<question>} placeholder is filled with one question sampled from the four candidates. Finally, the grammatical error correction prompt is applied during the evaluation phase, where it is used to assess the fluency of the generated answers. In this case, the \texttt{<text>} placeholder corresponds to the output produced by the VLM that is evaluated.

\begin{table}[htbp]
\centering
\small
\renewcommand{\arraystretch}{1.3}
\caption{Prompt templates used for question generation, answer generation, and grammatical error correction.}
\begin{tabular}{p{2cm} p{6.5cm} p{6.5cm}}
\hline
\textbf{Task} & \textbf{Romanian Prompt} & \textbf{English Prompt Translation} \\
\hline

\textbf{Question Generation} &
Ești un asistent vizual care primește o descriere de scenă în limba română. Generează patru întrebări clare și precise despre elementele scenei (obiecte, cantități, acțiuni și relații spațiale), asigurându-te că toate pot fi răspunse folosind exclusiv informațiile din descriere. Returnează întrebările sub forma unei liste structurate. 
\newline\newline
Descriere imagine: \textit{<descrierea imaginii>} 
&
You are a visual assistant that receives a scene description in Romanian. Generate four clear and precise questions about the elements of the scene (objects, quantities, actions, and spatial relations), ensuring that all questions can be answered strictly using the information provided in the description. Return the questions as a structured list.
\newline\newline
Image description: \textit{<image description>} \\

\hline

\textbf{Answer Generation} &
Ești un asistent vizual informat. Primești o descriere de scenă în limba română și o întrebare referitoare la acea descriere. Oferă un răspuns scurt (o singură propoziție), clar și direct, utilizând exclusiv informațiile furnizate în descriere.
\newline\newline
Descriere imagine: \textit{<descrierea imaginii>}
\newline
Întrebare: \textit{<întrebarea>}
&
You are an informed visual assistant. You receive a scene description in Romanian and a question related to that description. Provide a short (single-sentence), clear, and direct answer, using exclusively the information provided in the description.
\newline\newline
Image description: \textit{<image description>}
\newline
Question: \textit{<question>} \\

\hline

\textbf{Grammatical Error Correction} &
Ești un asistent AI pentru corectarea textelor în limba română. Primești un text, îl corectezi din punct de vedere ortografic, gramatical, al punctuației și al stilului, păstrând sensul original, și returnezi doar varianta corectată.
\newline\newline
Text: \textit{<text>}
&
You are an AI assistant for Romanian text correction. You receive a text and correct it for spelling, grammar, punctuation, and style, while preserving the original meaning. Return only the corrected version of the text.
\newline\newline
Text: \textit{<text>} \\

\hline
\end{tabular}
\label{tab:prompt_templates}
\end{table}